\documentclass[conference]{IEEEtran}
\IEEEoverridecommandlockouts
\usepackage{cite}
\usepackage{amsmath,amssymb,amsfonts}
\usepackage{algorithmic}
\usepackage{graphicx}
\usepackage{textcomp}
\usepackage{xcolor}
\usepackage{csquotes}
\usepackage{tikz}    
\usepackage{hyperref}

\newcommand{\mathpunkt}{\quad .}

\def\BibTeX{{\rm B\kern-.05em{\sc i\kern-.025em b}\kern-.08em
    T\kern-.1667em\lower.7ex\hbox{E}\kern-.125emX}}
\begin{document}

\title{Deep Neural Network Compression \\ for Image Classification and Object Detection\\
}


\author{Georgios Tzelepis$^{*,1}$, Ahraz Asif$^{*,1}$, Saimir Baci$^{1}$,  Selcuk Cavdar$^{1}$, and Eren Erdal Aksoy$^{1,2}$
\thanks{The research leading to these results has received funding from the Knowledge Foundation and the Vinnova FFI project SHARPEN, under grant agreement no. 2018-05001.}
\thanks{$^{1}$Volvo Technology AB, VGTT, Gothenburg, Sweden}
\thanks{$^{2}$Halmstad University, School of Inf. Technology, Halmstad, Sweden}
\thanks{$^{*}$The first two authors contributed equally to this work} 
}


\maketitle

\begin{abstract}
Neural networks have been notorious for being computational expensive. This is mainly because neural networks are often over-parametrized and most likely have redundant nodes or layers as they are getting deeper and wider. Their demand for hardware resources prohibits their extensive use in embedded devices and puts restrictions on tasks like real time image classification or object detection. 
In this work, we propose a \textit{network agnostic}  model compression method infused with a novel dynamical clustering approach to reduce the computational cost and memory footprint of deep neural networks.
We evaluated our new compression method on five different state-of-the-art image classification and object detection networks. 
In classification networks, we pruned about $95\%$ of network parameters. 
In advanced detection networks such as YOLOv3, our proposed compression method managed to reduce the model parameters up to $59.70\%$ which yielded $110\times$ less memory without sacrificing much  in accuracy. 

\end{abstract}

\section{Introduction}

Deep learning has shown remarkable performance in various application areas~\cite{AlexNet,Rothfuss18,SilverHuangEtAl16nature}.
An intriguing application of deep learning is within the field of autonomous driving, where neural networks are used in scene segmentation \cite{hemaskrcnn}, object detection \cite{hadsell2008deep} and route planning \cite{planning2017}.
Although neural networks are considered mature and reliable, they have the distinct disadvantage of being computationally expensive to train and use. In computationally resource constrained environments such as embedded systems for autonomous vehicles, it becomes paramount to optimize the networks with respect to both memory and inference time.
Thus, in domains where achieving high levels of accuracy is vital (e.g. due to safety reasons) and where resources are limited, methodologies to make neural networks more compact and efficient are essential. 

In order to leverage the use of state-of-the-art network architectures and computationally expensive techniques such as residual blocks or convolutional layers within resource constrained   embedded systems, it is important that investigations are conducted into how network architectures can be made more efficient through the reduction of parameters that have negative or no impact towards useful predictions. 
Furthermore, increasing computational efficiency results in a more energy-efficient neural network (as fewer computations have to be done  to achieve state-of-the-art results) and a more memory-efficient neural network by reducing requirements on storage.
Therefore, reduction in inference time can enable usage of, for instance, advanced image classification and object detection network architectures in \blockquote{real-time}. 
In addition, reduction in memory footprint can enable the deployment of network architectures to resource-constrained embedded environments where disk space or RAM is limited.

In this work, we introduce  a generic model compression method that is \textit{network agnostic} (i.e. not dependent on a specific neural network architecture) and has minimal impacts on accuracy, while reducing the inference time and memory footprint of a network.
Our proposed method has two stages: pruning and quantization. In the pruning phase, low-importance parameters are first removed and then the network is retrained  for a small number of epochs. This first stage is repeated until reaching
 a threshold accuracy drop. The following quantization stage rather projects  network parameters into forms that enable more computationally efficient arithmetic by, for instance, migrating from float  to integer representation.    
These two consecutive compression operations reduce the number of required computations while also lowering the memory footprint of deep networks.

We experimentally evaluated our model compression method on  different image classification (using MNIST \cite{mnist} and Cifar-10 \cite{CIFAR10} datasets as well as ResNet50\cite{he2015} model) and object detection tasks (using YOLOv3 \cite{redmon2018yolov3} and Faster-RCNN \cite{ren2015faster} networks). 
%
In classification networks, we reached up to $95\%$ pruning of network parameters.  In relatively more complicated detection networks, we managed to reduce the number of model parameters up to $59.70\%$ without sacrificing much  in accuracy.
With our proposed method, we achieved up to $182\times$ and $110\times$  memory compressions (i.e. reduction rates in memory required for network parameters) for the classification and detection networks, respectively.

Our main contributions in this work are as follows:

\begin{itemize}
    \item We applied model pruning to object detection neural network architectures. To the best of our knowledge, pruning methodologies have not been applied to object detection  networks. We showed that our proposed pruning method is capable of removing low-importance parameters for object detection while maintaining reasonable performance. 
    
    \item We extended the model compression methodology described by Han et. al. in \cite{han2015deep} with a dynamic method which yields better clustering of network weights during the process of k-means quantization. 
   We experimentally showed that using a dynamic method of clustering which determines the number of clusters required per layer in a deterministic fashion is capable of reducing the accuracy drop that is caused by k-means quantization.
      
\end{itemize}

\section{Related Work}
 
\textit{Model compression} refers to the techniques that reduce the inference time and the memory footprint of a deep network. Such compression methods generally leverage the use of network \textit{pruning} and \textit{quantization}.

\subsection{Pruning}

Network pruning involves reducing the number of parameters of a network (e.g. weights) through their removal based on a criterion or threshold. There exist several proven methods of ranking network parameters based on a variety of different criterion, from generating thresholds based on the distribution of the weights~\cite{han2015deep} to more sophisticated parameter ranking techniques such as using Taylor Expansion to approximate the importance of a parameter~\cite{molchanov2016}.

Molchanov et. al. \cite{molchanov2016} compared magnitude-based pruning criterion in~\cite{han2015deep} against a novel criterion derived by using a first order Taylor polynomial to approximate the change that removing a parameter will have on the network cost function. The greater the impact, the more important a parameter is predicted to be. In their results, the Taylor Expansion criterion outperformed the magnitude based pruning on AlexNet trained on the ImageNet dataset. 

Zhuang et. al. \cite{zhuang2017} pruned convolutional layers by using the scaling factor learned by Batch Normalization layers for channels to estimate their importance, of which a specified percentage of those channels are pruned. 
A greedy pruning algorithm was defined in \cite{luo2017} to remove a channel based on its impact on the activation values of the next layer (i.e. the one with the smallest impact is pruned). This method performed better than using Taylor Expansion (in terms of accuracy) to determine the importance of a network parameter on the VGG-16 network architecture.

In \cite{see2016}, See et. al. investigated the use of three different model pruning techniques, with the goal of reducing the number of weights of neural machine translation architectures: class-distribution, class-uniform and class-blind. They define the concept of a \blockquote{weight class}, subsets of all of the weights of the model. Essentially, these weight classes can be considered as the weights per layer, i.e. the input weights of a given layer would all be the same weight class.

Class-distribution pruning, as employed in \cite{han2015deep}, defines a threshold value based on the standard deviation of the given weight class, controlled by a scaling hyperparameter. All weights with a magnitude lower than the threshold value are removed from the network, reducing memory and inference time. Furthermore, the class-uniform method of pruning removes the same percentage of weights from each of the weight classes, such that the total number of parameters pruned is equal to the number of parameters pruned per weight class. Finally, class-blind pruning does not consider weight class at all; instead, all of the weights across the network are ordered by magnitude and the bottom percentage, as controlled by a hyperparameter, are pruned (e.g. the bottom 20\% of the weights for a network).

See et. al. \cite{see2016} found the class-blind method to be the most effective in regards to the percentage pruned versus the accuracy of the pruned model. Class-blind pruning drops off the least in regards to accuracy, with class-distribution and class-uniform pruning both dropping at a sharper rate after around 30\% of parameters pruned. The intuition here being that as the class-blind method compares the weights globally, it is more biased towards removing low magnitude weights than class-distribution or class-unfirom weights, which will prune a significant proportion of all weight classes regardless of their overall magnitude.
 
A common practise after pruning is to fine-tune the remaining weights by retraining the network to improve accuracy. Pruning has often acted as a form of regularization by actually increasing the accuracy of the model as shown in \cite{see2016} where  consecutive retraining and pruning of the network are performed until the accuracy begins to decrease. This strategy enables the network to retain as few weights as possible. 



\subsection{Quantization}

Quantization refers to the act of reducing the memory required to represent a network parameter \cite{guo2018}. Existing quantization techniques in the literature fall into two large categories: deterministic and stochastic quantizations. Deterministic quantization involves mapping a network parameter to a value such that the parameters occupy less space. Stochastic quantization models network parameters as discrete distributions and samples from them.
 
Rounding is an example of a deterministic quantization technique. A parameter is mapped directly to a quantized parameter that occupies less space, typically at a lower precision. For example, rounding a weight down from float64 to float32 can be seen as a form of deterministic quantization. There is a loss in accuracy of the model depending on how severe the rounding operation is. Furthermore, rounding can also destroy the training by reducing the precision of the gradients (i.e. flattening them). Therefore, many deterministic quantization algorithms keep track of the original weight value during training, and subsequently re-quantizing.
 
The work in  \cite{courbariaux2015} introduced a rounding-based quantization by quantizing weights into binarized weights of values:+1 (for positive weights) or -1 (for negative weights). Even with such an extreme rounding strategy they were able to outperform bespoke baseline classifiers on CIFAR-10, MNIST and several other datasets.
Polino et. al. \cite{polino2018} proposed an alternative rounding quantization technique that employs a scaling function  to map a range of arbitrary values. 
This method reduced network memory footprint by up to an order of magnitude before seeing a significant drop in accuracy.

Vector quantization is another deterministic method. It involves clustering weights using traditional clustering techniques (e.g. k-means) and using the centroids as the weight values. By mapping all the weights to a respective centroid, the total amount of weight values to be stored is reduced, as a low-bit index to the centroid can be used instead. 
In \cite{han2015deep}, the authors used  k-means clustering on a per-layer basis to generate centroids for the weights. During training, gradients were aggregated per centroid and simply summed and added to the centroid. This enabled the centroid to benefit from training without destroying the accuracy for the entire model. 

In stochastic quantization, network parameters such as weights or gradients are rather modeled as discrete distributions, and instead sampled from a distribution. One such example of this is when the parameters of these distributions are typically inferred through the use of learning algorithms such as the Expectation Back-propagation algorithm \cite{soudry2014}. Furthermore, the use of stochastic quantization also adds a regularization effect to the model. Finally, only the distribution and its parameters have to be stored thus reducing the memory occupied by representing the parameters.

This is referred to as probabilistic quantization and can be highly efficient as shown in \cite{soudry2014} where the authors leveraged mean-field approximation and the Central Limit Theorem   to approximate the posterior distribution of the network weights (an intractable problem). Their results showed that the Expectation Back-propagation algorithm outperformed standard back-propagation.
 However, such techniques are not applicable to all network architectures, such as recurrent neural networks.

\section{Method}\label{method}

\begin{figure}[!b]
    \centering
    \includegraphics[scale=0.55]{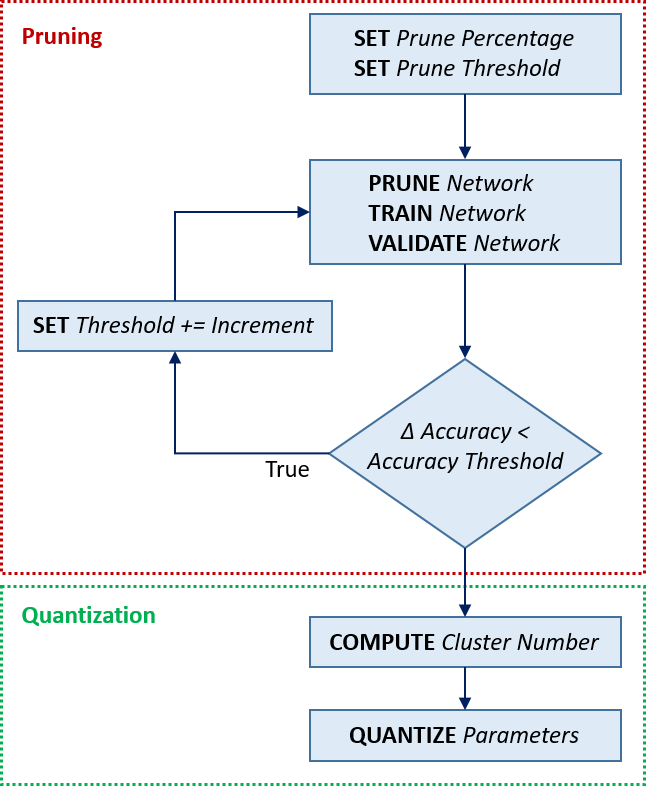}
    \caption{Our proposed two-stage network compression method. }
    \label{fig:prunePipeline}
\end{figure}

Fig.~\ref{fig:prunePipeline} depicts the pipeline of our proposed method which starts with iterative class-blind pruning (shown in red dashed box) and then continues with the k-means quantization of  network parameters (green dashed box).

\subsection{Pruning} \label{par:pruningmethodology}
In this stage, we initially assign a percentage of parameters to be pruned. This value was then employed to derive a prune threshold which specifies the boundary value for parameters to be pruned. Practically, any network weight with a magnitude below this threshold was pruned. Next, the model was pruned using the threshold value, re-trained using a small number of epochs (between three and five) and set of weights producing the highest accuracy between the epochs were maintained. Finally, the network was validated using a test data set. If the accuracy was seen to have dropped beyond that of the specified accuracy threshold, the pruning was ceased and the model was saved. However, if the accuracy threshold had not been reached, the prune percentage was incremented by a small amount to continue removing more network weights.

\subsection{Quantization} \label{par:quantizationmethodology}

After pruning, network parameters were quantized by using k-means quantization. 
Unlike the original k-means quantization method proposed in \cite{han2015deep}, we dynamically compute the number of clusters supplied to the k-means algorithm while sorting the weights of a given layer.
When the number of clusters were too low, especially for networks with a large number of weights such as YOLOv3, there was a significant decrease in the accuracy of the network. However when the number of clusters were too high, the quantization algorithm was almost unusable due to the huge amount of memory needed. Therefore, we introduce a new deterministic equation  to dynamically scale the number of clusters based on the number of parameters within a given layer, as formulated in Eq. (\ref{numberofclusters}).

\begin{table}[!t]
\caption{mAP of YOLOv3 on COCO2014 with and without our proposed dynamic k-means quantization} 
\begin{center}
\begin{tabular}{|l|l|l|}
\hline
\textbf{Initial} & \textbf{Quantized with}  & \textbf{Quantized with our} \\
                 & \textbf{k=32 clusters}   & \textbf{dynamic k-means clustering} \\ \hline
0.5890           & 0.4400                   & 0.5010 \\ \hline
\end{tabular}
\label{tab:dynamicclustering}
\end{center}
\end{table}

Let $\mathcal{C}$ and $N$ denote the number of clusters and parameters per set of clusters respectively. Then for each layer $l$  with $P$ number of parameters, the number of clusters $c$ is given by
\begin{align}\label{numberofclusters}
    c_l = \left \lceil{\frac{P_l}{N_l}}\right \rceil * \mathcal{C}_l 
    \mathpunkt
\end{align}{}

Unlike the static quantization method, such as the one introduced in \cite{han2015deep}  where 32 clusters where generated for every layer, our proposed dynamic clustering depends only on the number of parameters in each layer. 
This way, the quantized layer will be closer to the original distribution of the weights, which is particular beneficial since the parameter number in the later layers of a deep network is large.

The impacts of the use of this dynamic method of clustering can directly be seen in Table \ref{tab:dynamicclustering}. By using a greater number of clusters (e.g. 32 clusters) for larger layers, the decrease in mean Average Precision (mAP) on YOLOv3 on the COCO2014 dataset was minimized, i.e. the mAP dropped to $0.5010$ instead of $0.4400$ from $0.5890$.

Unlike the quantization method in \cite{han2015deep}, we also map the cluster centroid back over the clustered weights instead of utilizing a code-book. Code-books are mainly employed to map the weights of clusters to a low-bit identifier to be later used for a lookup to the quantized weight value. The main advantage coming with our naive mapping approach is the speed increment in the process of searching for the corresponding weights in the code-book.

\begin{figure}[!t]
    \centering
    \includegraphics[scale=0.45]{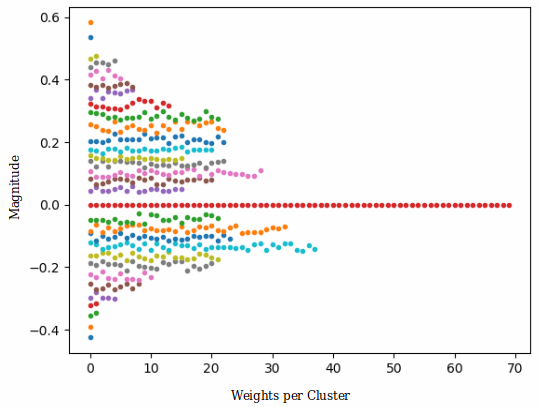}
    \caption{K-means Weight Quantization on MNIST Classifier}
    \label{fig:kmeansquant}
\end{figure}

The effects of k-means quantization \textit{after pruning} on an early layer of the MNIST Classifier architecture is visualized in Fig.~\ref{fig:kmeansquant}. The data points represent individual weight values grouped by cluster and each data point was colored with its respective cluster. As expected, there was large cluster of values with a magnitude of zero, due to the sparsification caused by pruning. The clusters were spaced equidistantly between the minimum and maximum magnitude of the available weights. This enabled weights with extreme magnitudes to not be pulled into the nearest cluster, which could potentially cause a large change in the value. The centroid for each weight cluster was mapped back over the weight to simulate the effects of k-means quantization.

\section{Experiments}\label{experiments}

For an experimental evaluation, we applied our proposed compression method to several state-of-the-art neural network architectures within the domains of image classification and object detection problems, with the aim  to optimize the inference time and memory footprint of  networks with different depth and width. For the classification task, we used MNIST \cite{mnist} and Cifar-10 \cite{CIFAR10} datasets and ResNet50\cite{he2015} model, whereas more complex networks such as YOLOv3 \cite{redmon2018yolov3} and Faster-RCNN \cite{ren2015faster} were employed for the object detection task. 
Large-scale, public and open datasets were chosen for evaluation using open-source implementations of popular neural network architectures.  For all the assessed architectures, pre-published pre-trained weights  were used as a baseline. All implementations of our proposed compression method were done in PyTorch and CUDA and are publicly available 
\footnote{\href{https://github.com/AhrazA/modelcompression-2019}{{https://github.com/AhrazA/modelcompression-2019}}}.

\subsection{MNIST Classifier}\label{MNIST_clas} 

The first architecture used for evaluation was a classifier for the MNIST dataset \cite{mnist}. It consisted of a convolutional layer with one input channel, 20 output channels and a 5x5 kernel. The next layer was another convolutional layer with 20 input channels, 50 output channels and a 5x5 kernel. Subsequently there were two fully connected layers with 800 input channels and 500 output channels, and finally with 500 input channels and 10 output channels. All of the layers used ReLU activation function and max pooling was applied after each layer. After the final layer a softmax function was applied to obtain the class probabilities.

\subsection{Cifar  Classifier} 

Unlike the previous shallow MNIST classifier, we constructed a deeper classifier for the  Cifar-10 \cite{CIFAR10} images dataset. This deep network consisted of two parts: feature extractor and classifier. The feature extractor stage had three consecutive convolutional blocks which included batch normalization and dropout before downsampling with max pooling. The classifier part was a fully-convolutional block with basic residual network units. It is worth nothing that dropout should be in place after batch normalization, otherwise, as noted in \cite{li2018understanding}, it can cause a shift in the weight distribution and thus minimizing the effect of batch normalization during training.

\subsection{ResNet50}  

Residual Networks\cite{he2015} were implemented for the Stanford Dogbreeds dataset, which is a subset of Imagenet \cite{imagenet_cvpr09} data focusing on dogs, to classify 120 classes of dogs. Residual Networks were made consisting of blocks  that use skip connections to transfer the input information after to the last layer of the block. Also batch normalization was applied after each convolution in order to help converging to the optimal solution by solving the internal covariate shift. More specifically the network to be used was ResNet50.

\subsection{YOLOv3}

We further employed a state-of-the-art object detection network architecture YOLOv3 \cite{redmon2018yolov3} which has been shown to have astonishing performance in object detection problems regarding the inference time. It is a one-stage detector where the fully convolutional network predicts the bounding boxes and class probability of those boxes in one pass.

The network is made up of 75 convolutional layers with no fully connected layers. In addition the architecture is inspired by Residual Networks (ResNet) and from Encoder-Decoder models such as Feature Pyramid Networks (FPN)\cite{lin2017feature}. The encoder, which is responsible for down-sampling and feature extraction, has a stride of two to be used instead of max-pooling layers. Subsequently an FPN is used to maintain the details that are lost during the encoding part of the network by applying element-wise summation to the corresponding layers of the decoder, while ResNet adds skip connections within the residual blocks that allow the network to be influenced from previous layers in order to generate better features.

\subsection{Faster-RCNN}

Another state-of-the-art object detection algorithm is a two-stage detector Faster-RCNN \cite{ren2015faster}. The network takes the last feature map of a convolutional layer and by using an operation called Regional Proposal Networks (RPNs), it separates what is likely background and what likely an object. The areas that are candidates containing an object are called Regions of Interest (RoI) and are the ones that are used to detect the objects through further processing by the detection network.

Faster-RCNN is composed of three different neural networks; (1) The Feature Network which is responsible for constructing the features and is usually a ResNet, (2) The RPN that is usually a plain three layered convolutional network which detects the regions on the last feature map with a high probability of containing an object and generates a number of bounding boxes for them called RoI and (3) the Detection Network, that is usually made up of four Fully Connected layers which take those RoIs produced by the RPN and generate the final class and bounding box.

\section{Results}\label{results}

Table \ref{table:compressionresults} shows the performances of five different model architectures, three image classifiers and two object detectors described in section~\ref{experiments}, at different stages of our model compression pipeline depicted in Fig. \ref{fig:prunePipeline}. The column named \textit{Initial} presents the accuracy or mAP of the architecture using pre-trained weights before our model compression pipeline is applied. The \textit{Pruned} column indicates the accuracy or mAP after the pruning step of our pipeline. The \textit{Pruned \& Quantized} column presents the final accuracy or mAP after the entire model compression procedure has been applied. The very last column, called \textit{Percentage Pruned}, indicates the total  percentage of the network parameters that were removed. 

\begin{table}[!b]
\begin{center}
\caption{Model Compression Results}
\scalebox{0.88}{
\begin{tabular}{|l|c|c|c|c|}
\hline
\textbf{Architecture (metric)}  & \textbf{Initial} & \textbf{Pruned} & \vtop{\hbox{\strut \textbf{Pruned \&}}\hbox{\strut \textbf{Quantized}}} & \vtop{\hbox{\strut \textbf{Percentage}}\hbox{\strut \textbf{~~Pruned}}} \\ \hline
MNIST Classifier (accuracy)     & 0.990    & 0.993            & 0.993               & 95.00        \\ \hline
CIFAR10 Classifier (accuracy)   & 0.917    & 0.899            & 0.898               & 95.00        \\ \hline
ResNet50 Classifier (accuracy)  & 0.834    & 0.825            & 0.747               & 45.00        \\ \hline
YOLOv3 Detector (mAP)           & 0.589    & 0.537            & 0.530               & 59.70        \\ \hline
FasterRCNN Detector (mAP)       & 0.677    & 0.662            & 0.612               & 52.30          \\ \hline
\end{tabular}
\label{table:compressionresults}
}
\end{center}
\end{table}

Furthermore, presented in Table \ref{table:compressionmemoryresults} are the predicted memory requirements for each tested network at different stages. All network parameters are assumed to be float32 tensors occupying four bytes. For the results presented at the \textit{Pruned} step, all the weights that have been removed from the model are excluded. For the \textit{Pruned \& Quantized} step, the compression ratio for the parameters remaining is calculated using Eq. (\ref{remainingparameter}) that was initially introduced in \cite{han2015deep}. Per layer, given $k$ clusters, $log_{2}(k)$ bits are required to encode the indices for the references to the cluster centroids. Given $n$ weights, with each connection being represented by $b$ bits and given that only $k$ full precision weight values will be stored, the compression rate $r$ is given by:

\begin{align}\label{remainingparameter}
    r = \frac{nb}{nlog_{2}(k) + kb}
    \mathpunkt
\end{align}{}

Using this compression ratio, the number of bytes occupied by the quantized model was estimated.

\begin{table}[!t]
\begin{center}
\caption{Predicted Memory Reduction Results}
\scalebox{0.9}{
\begin{tabular}{|l|r|r|r|c|}
\hline
                      & \multicolumn{4}{c|}{\textbf{\begin{tabular}[c]{@{}c@{}}Memory required for\\ network parameters (bytes)\end{tabular}}}                                                  \\ \hline
\textbf{Architecture} & \multicolumn{1}{l|}{\textbf{Initial}} & \multicolumn{1}{l|}{\textbf{Pruned}} & \multicolumn{1}{l|}{\textbf{\begin{tabular}[c]{@{}l@{}}Pruned \&\\ Quantized\end{tabular}}} & \multicolumn{1}{l|}{\textbf{\begin{tabular}[c]{@{}c@{}}Compress\\ Rate\end{tabular}}}\\ \hline
MNIST Classifier      & 1724320                & 88420                  & 9478           & 182$\times$           \\ \hline
CIFAR10 Classifier    & 23412032               & 15239888               & 2390274        & 9.8$\times$             \\ \hline
ResNet50 Classifier   & 95228552               & 63072340               & 1584860        & 60$\times$             \\ \hline
YOLOv3  Detector      & 247586164              & 99036300               & 2239175        & 110$\times$           \\ \hline
FasterRCNN  Detector  & 190270652              & 140390612              & 6472874        & ~29$\times$            \\ \hline
\end{tabular}
\label{table:compressionmemoryresults}
}
\end{center}
\end{table}

Both the image classification and object detection architectures were pruned and quantized using the aforementioned compression pipeline (see Fig. \ref{fig:prunePipeline}).
As indicated in Table \ref{table:compressionmemoryresults}, after these pruning and quantization steps we reach up to 182$\times$ compression rate for the MNIST classifier and 110$\times$ for the YOLOv3 detector.

Presented in the following subsections are the percentage of parameters (weights) pruned and the accuracy after the retraining step had completed. 
As a baseline, we used the layer-wise pruning from \cite{han2015deep} and compared with our class-blind pruning method for each network below.

\subsection{MNIST Classifier}

\begin{figure}[!t]    
  \centering
  \begin{tikzpicture}
  \node(picA){\includegraphics[scale=0.65]{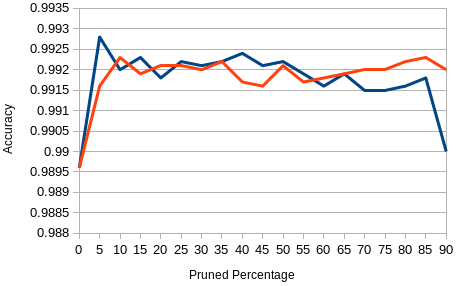}};
  \node at (1,-1 ){\includegraphics[scale=0.65]{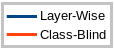}};
  \end{tikzpicture} 
  \caption{MNIST Classifier Pruning Results}
  \label{fig:mnistprune}   
\end{figure}

The dataset used for training and evaluation was the MNIST dataset, a popular hand-writing classification dataset \cite{mnist}. As shown in Fig. \ref{fig:mnistprune} the MNIST classifier architecture exhibited very little improvement in accuracy through pruning. With our class-blind method, the MNIST classifier actually began to increase in accuracy, all the way up to 95\% of parameters pruned, whereas this is the case only up to 40\% for baseline layer-wise pruning (see the blue curve in Fig. \ref{fig:mnistprune}).  With our pruning method, the final difference between the initial accuracy of the trained-from-scratch model and the pruned model at 95\% pruning was about $+0.003$, indicating a highly over-parameterized model. Furthermore, the increase in accuracy could be attributed to the additional training conducted on the sparse model, or could be a symptom of over-fitting to the training data. 
 
Note that, at a 95\% prune rate,  layer-wise pruning accuracy drops below 0.1, therefore, the plot in Fig. \ref{fig:mnistprune} is shown only up to 90\% pruning rate.
Also, at a 100\% prune rate the accuracy of our class-blind method drops to 0.1, as expected since there are ten classes to predict. This indicates that the model was not making any meaningful predictions at the 100\% prune rate. Therefore, the result at 100\% prune percentage was omitted from Fig. \ref{fig:mnistprune}. Furthermore, after quantizing even at 95\% pruning there was essentially no loss in accuracy, as seen in Table~\ref{table:compressionresults}.

\subsection{CIFAR Classifier}
\begin{figure}[!b]    
  \centering
  \begin{tikzpicture}
  \node(picA){\includegraphics[scale=0.65]{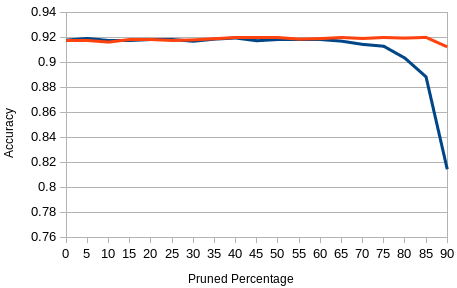}};
  \node at (1,-1 ){\includegraphics[scale=0.65]{legend.png}};
  \end{tikzpicture} 
  \caption{CIFAR Classifier Pruning Results}
  \label{fig:cifarprune}   
\end{figure}



The dataset used for training and evaluation was the CIFAR10 dataset consisting of 10 classes \cite{CIFAR10}. The CIFAR classifier experienced minor drop in performance when being pruned as shown in Fig. \ref{fig:cifarprune}. There is also a slight  initial increase in accuracy which remains until approximately 85\% of the parameters have been pruned.  Once again, this indicates a highly over-parameterized model of which the ideal network architecture can be much sparser. Finally, the CIFAR classifier had an accuracy drop of approximately 0.018 (i.e. 1.8\%) at a pruning percentage of 95\% (see  Table~\ref{table:compressionresults}). However, the accuracy drop in the layer-wise method started much earlier (about pruning percentage of 65\%) and continued dramatically afterwards. Note that due to the very same reason mentioned in the previous section, the plot in Fig. \ref{fig:cifarprune} is shown   up to 90\%.

The subsequent quantization step in our method introduced a small additional accuracy loss of 0.001 (see  Table~\ref{table:compressionresults}). At 100\% pruning the accuracy falls to 0.1, which indicates that the model is not making any meaningful predictions. Therefore the result at 100\% pruning was again omitted from Fig. \ref{fig:cifarprune}.

\begin{figure}[!t]    
  \centering
  \begin{tikzpicture}
  \node(picA){\includegraphics[scale=0.65]{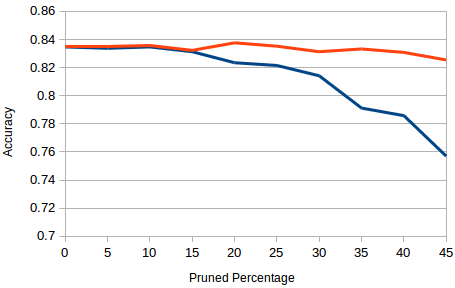}};
  \node at (1,-1 ){\includegraphics[scale=0.65]{legend.png}};
  \end{tikzpicture} 
  \caption{ResNet50 Classifier Pruning Results}
  \label{fig:resnet50prune}   
\end{figure}

\subsection{ResNet50}\label{resnet}  
As shown in red curve in Fig.~\ref{fig:resnet50prune}, we managed to prune 45\% of  ResNet50 model parameters while having a small accuracy drop from 0.834 to 0.825 (see also  Table~\ref{table:compressionresults}). Unlike our method, the baseline layer-wise pruning, i.e. blue curve in Fig.~\ref{fig:resnet50prune}, caused an accuracy drop at earlier phase (around 15\%) and had about 0.76 accuracy value at pruning percentage of 45\%. This is a clear indication of the success of our proposed class-blind pruning over the baseline method. 

Our final quantization stage introduced an additional drop and therefore the final accuracy value was 0.747 (see Table~\ref{table:compressionresults}).

\subsection{YOLOv3}\label{yoloprun}
 
\begin{figure}[!b]    
  \centering
  \begin{tikzpicture}
  \node(picA){\includegraphics[scale=0.65]{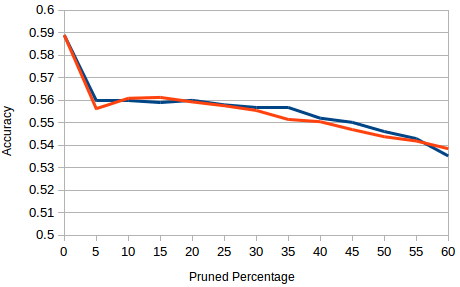}};
  \node at (1,-1 ){\includegraphics[scale=0.65]{legend.png}};
  \end{tikzpicture} 
  \caption{YOLOv3 Detector Pruning Results}
  \label{fig:yoloprune}   
\end{figure}



The dataset used for training and evaluation was the COCO2014 dataset \cite{lin2014microsoft}, a popular dataset for object detection. Having applied the prune pipeline to YOLOv3, as shown in Fig. \ref{fig:yoloprune}, with a prune threshold of 0.05, the model was pruned up to 59.70\% sparsity before the threshold was reached. The initial 5\% pruned caused the steepest decline in the mAP, indicating the potential loss of important weights with low magnitudes. However, subsequent pruning iterations led to small increases in accuracy before a slow decline until the 59.70\% mark. The large number of parameters pruned with a relatively small impact on the mAP indicates that YOLOv3 is overparameterised and sparsification can lead to a faster, smaller model. Furthermore, it is worth noting that after every pruning iteration the model was retrained for three epochs. It is possible that significantly increasing the number of retraining epochs might minimize the drop in mAP. Note that, unlike the previous models, our proposed method exhibited a comparable behaviour  with the baseline method as depicted in Fig. \ref{fig:yoloprune}.

When undergoing the full compression pipeline, a prune percentage of 59.70\% (see  Table~\ref{table:compressionresults}) was applied as it had the best weights for the given compression run. Subsequently, the quantization of the network resulted in a minimal decrease (0.7\%) in the mAP of the model. The overall loss in mAP through our entire model compression pipeline was 0.059 as indicated in Table~\ref{table:compressionresults}.

\begin{figure}[!b]    
  \centering
  \begin{tikzpicture}
  \node(picA){\includegraphics[scale=0.61]{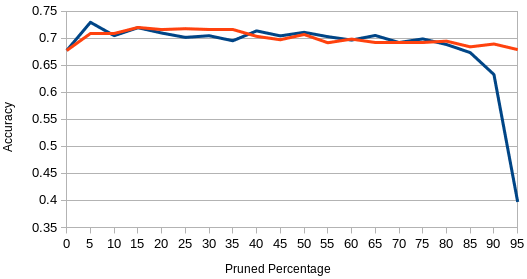}};
  \node at (1,-1 ){\includegraphics[scale=0.65]{legend.png}};
  \end{tikzpicture} 
  \caption{FasterRCNN Detector Pruning Results}
  \label{fig:fasterprune}   
\end{figure}

\subsection{FasterRCNN}\label{fasterprune}

The dataset used for training and evaluation was the VOC2007 dataset \cite{pascal-voc-2007}, another popular dataset for object detection. Within the FasterRCNN architecture only the base (detector) of the network was pruned, which represents approximately 52.70\% of the entire network weights. Fig. \ref{fig:fasterprune} indicates the percentage of parameters of \textit{ only the base of the network}.
Unlike YOLOv3, after pruning FasterRCNN there was much less drop (0.015 mAP) in accuracy. In fact, the iterative pruning methodology pruned the detector  parameters up to 100\%  before the initially specified prune thresholds (0.05) were reached. This indicates that the detector is heavily over-parameterised, with the network having a drop in mAP of 0.025 at 95\% parameters pruned (see Fig. \ref{fig:fasterprune}). The peak mAP reached was at 15\% parameters pruned with an increase in mAP of approximately 0.02. As the prune percentage was incremented units of 5\%, it is entirely possible that a finer interval could result in an even higher accuracy. At a pruning percentage of 100\% the mAP dropped to 0.1, which indicates that the model is not making any meaningful predictions as expected. Therefore, the result at 100\% pruning was omitted from Fig.~\ref{fig:fasterprune}. Unlike our proposed method, the layer-wise pruning caused a significant drop after 80\% pruning rate.


The subsequent quantization resulted in an overall decrease in mAP of 0.065 (see Table~\ref{table:compressionresults}). Consequently, the prune percentage of 52.30\% was applied as it resulted in the highest mAP during the compression run, and the quantization step caused the largest decrease in mAP.

\section{Discussion}\label{discussion}

In this study, we have shown how class blind pruning and quantization can reduce the size of the model. The main idea was to find the parameters over the whole model that have minimal impact and set them to zero, thus reducing the computational cost. Class blind pruning that is applied on the whole model was chosen instead of layer-wise pruning used in \cite{han2015deep} due to the fact that network tends to over parameterize in the deep layers. Thus by using class-blind pruning, it is possible to maintain the early parameters which are important for the construction of the high level features while removing the ones which are not contributing much into the final predictions. All our experimental results confirmed that class-blind pruning results in less drop in accuracy in contrast to layer-wise pruning.

Model Compression in the two object detection architectures chosen proved to be very promising, with the detector of the FasterRCNN being almost entirely prunable with minimal impact to accuracy. Furthermore, YOLOv3 was pruned to quite a significant extent with very little drop in mAP, with up to 50\% of its parameters pruned while maintaining not breaking the threshold of 0.05 mAP. This indicates that the use of our model compression method in advanced architectures, such as these deep object detectors, can enable execution on resource constrained environments in real-time scenarios.

The percentage of the weights that were pruned were depending largely on the model, the dataset and the task that network was performing. Object Detectors despite the fact that they have vastly more parameters than classifiers, arrived at a lower pruning percentage. Even for Faster-RCNN which was using ResNet101 as a backbone that contains 44.5 million parameters and was trained on a realatively easy dataset as VOC2007, the percentage of pruned weights was lower than all the classifiers. 

The use of pruning and quantization can be seen as managing a tradeoff between performance and accuracy; the greedier the pruning (i.e. high prune percentages) and quantization strategies applied (i.e. quantization into fewer clusters or lower bits) the greater the loss in performance at the expense of reduced computation. The application of such methodologies should attempt to address the usage context of the generated network before settling on how much the models should be compressed.

Class-blind pruning often caused a performance increase at different percentages when a model is vastly overparametarized. This regularization effect is likely due to the removal of low-importance weights that simply interfere with network predictions, however often the percentage of parameters pruned were very high when the performance increase manifested. A similar effect has been shown  with targeted dropout\cite{gomez2018targeted} where dropout is applied on the lower absolute  values of the layers. 
This regularization effect, i.e. increase in the prediction accuracy after pruning, has been observed in Faster-RCNN. However that is not the case with YOLOv3 where an initial drop occurs.
We believe that the main reason why this accuracy increase does not occur in YOLOv3 is the fact that it was trained on MS-COCO which is a rather difficult dataset with more classes than VOC2007 and more objects with similar features to each other. 
Consequently, as our experimental findings suggest, the regularization theory as stated by Han et. al. \cite{han2015deep} appears to hold true even for high-complexity models such as FasterRCNN.

Han et. al. demonstrated in \cite{han2015deep} that the sparsification (i.e. pruning) of weight matrices leads directly to lower inference times. This is because fewer floating point operations have to be conducted and fewer gradients having to be computed due to fewer weights being involved with inference. However, due to limitations with PyTorch and CUDA regarding the use of sparse matrices, we were unable to directly collect the inference time improvements as a result of pruning. However it still holds true that, for example, pruning 50\% of network parameters will lead to a proportionate 50\% drop in inference time for forward and backward propagation, as shown in \cite{han2015deep}. 
As an extension of this work, we are planning to employ an alternative framework, such as TensorFlow where sparse matrices are available, in order to solidify this claim.

For the Quantization choosing k-means as a method of clustering was implemented to run on the GPU since, applying k-means over millions of parameters was not efficient. The identity in Eq.~(\ref{numberofclusters}) was utilized in order to create a dynamic clustering that was dependant on the number of parameters for each layer. While the weights were clustered significantly there was not a significant drop in the accuracy. However at least a small drop in the accuracy is to be naturally expected by using quantization which is more to the fact that it requires clustering, than lowering the precision of the weights.

\section{Conclusion}\label{conclusion}

We presented a model compression method to reduce the inference time and memory footprint of advanced deep neural networks.
We showed the application of our method on different image classification and object detection models.

We presented various experimental findings showing the predicted memory footprint impacts of k-means quantization on the network architectures that the pruning pipeline was applied to.
Our findings on reduction of memory footprint are very much aligned  with the results in Han et. al. in \cite{han2015deep} which concretely showed the memory footprint impacts that k-means quantization can have. 

We note that reduction of inference time was supported by the findings presented in \cite{han2015deep}, showing that sparsification of weights matrices can speed up inference time (and provide more energy efficient networks) by the reduction in floating point operations that are to be conducted in a forward or backward pass. While the investigation presented in this study was unable to obtain concrete inference time measurements due to lack of capabilities of software tools, it was shown that the inference time speedups are theoretically possible.

 
\bibliographystyle{IEEEtran}
\bibliography{ModelCompBib}


\end{document}